\documentclass[a4paper,11pt]{article}

\usepackage[margin=1in]{geometry}
\usepackage[T1]{fontenc}
\usepackage[utf8]{inputenc}
\usepackage{newtxtext,newtxmath}
\usepackage{setspace}
\usepackage{microtype}
\usepackage{amsmath}
\usepackage{graphicx}
\usepackage{booktabs}
\usepackage{array}
\usepackage{tabularx}
\usepackage{pdflscape}
\usepackage{float}
\usepackage{placeins}
\usepackage{flafter}
\usepackage{caption}
\usepackage[nocompress]{cite}
\usepackage{xcolor}
\usepackage{hyperref}
\hypersetup{colorlinks=true, linkcolor=blue!50!black, citecolor=blue!50!black, urlcolor=blue!50!black}
\graphicspath{{./}{figures/}}
\newcommand{\figref}[1]{\hyperref[#1]{Fig.~\ref*{#1}}}
\newcommand{\tabref}[1]{\hyperref[#1]{Table~\ref*{#1}}}

\makeatletter
\renewcommand{\maketitle}{%
  \begin{center}
  {\LARGE\bfseries \@title\par}
  \vspace{1.0em}
  {\normalsize \@author\par}
  \vspace{0.8em}
  \end{center}
}
\makeatother

\title{Active rejection enables reliable generalization of universal machine-learning interatomic potentials}

\author{
Mingxiang Luo\textsuperscript{1,2\textdagger}
\quad
Xinnan Mao\textsuperscript{4\textdagger}
\quad
Lu Wang\textsuperscript{4}
\\
Lei Bai\textsuperscript{3}
\quad
Feng Ding\textsuperscript{4*}
\quad
Yuqiang Li\textsuperscript{3,2*}
\\[0.8em]
\small \textsuperscript{1}School of Future Information Innovation, Fudan University, Shanghai, China
\\
\small \textsuperscript{2}Shanghai Innovation Institute, Shanghai, China
\\
\small \textsuperscript{3}Shanghai Artificial Intelligence Laboratory, Shanghai, China
\\
\small \textsuperscript{4}Suzhou Laboratory, Suzhou, China
}

\date{}

\begin{document}

\maketitle

\begin{abstract}
Universal machine learning interatomic potentials (uMLIPs) have revolutionized atomistic simulations by bridging quantum-mechanical accuracy with large-scale molecular dynamics. The prohibitive cost of high-accuracy calculations, such as r$^2$SCAN, limits high-fidelity uMLIP training to datasets that are much smaller and less diverse than the open materials space. Models trained under this constraint can exhibit localized reliability gaps, where strong benchmark performance does not ensure reliable energy--force predictions for every structure. A practical route is to mine massive medium- to low-fidelity structure repositories with multiple pretrained uMLIPs while filtering unreliable pseudo-labels. Here, we propose the Adaptive Multi-Teacher Routing (ATR) framework, which reformulates high-fidelity data construction as a structure-wise decision problem under uncertainty. Using a small set of real r$^2$SCAN labels, ATR calibrates multiple pretrained uMLIP teachers and combines structural descriptors, teacher identity and inter-teacher disagreement to estimate the reliability of each structure--teacher pair. It selects high-confidence teacher predictions for pseudo-label generation and rejects structures for which no teacher is sufficiently reliable. With real r$^2$SCAN labels for only 0.2\% of the candidate structures, ATR distils 2.89 million traceable r$^2$SCAN-level pseudo-labels for model pretraining. Experiments on held-out r$^2$SCAN structures and the MP-r$^2$SCAN benchmark show that a lightweight CHGNet trained on the ATR-generated dataset consistently outperforms the baseline and non-routed controls. Finite-temperature molecular dynamics validations further show that ATR pretraining improves dynamical robustness across multiple material systems, maintaining stable trajectories in cases where baseline simulations undergo catastrophic structural collapse. These results establish active rejection as an effective mechanism for converting multiple pretrained uMLIPs into a scalable and reliable data-construction system, improving the generalization of high-fidelity uMLIPs across the open materials space.
\end{abstract}

\section{Introduction}

Machine learning interatomic potentials (MLIPs)\cite{Deringer2019MLIPReview} are substantially expanding the accessible time scales, length scales and application scope of atomistic materials simulations\cite{Zhang2018DPMD,Wang2024AI2BMD,Bian2024SlidingFerroelectrics}. Large-scale electronic-structure databases and increasingly expressive graph neural networks\cite{Schutt2018SchNet,Schutt2021PaiNN,Batzner2022NequIP,Musaelian2023Allegro} have enabled universal MLIPs (uMLIPs)\cite{Chen2022M3GNet,Takamoto2022PFP,Deng2023CHGNet} that support structure optimization, molecular dynamics, phase-stability assessment and property prediction across diverse materials spaces. In this sense, uMLIPs are becoming atomistic scientific foundation models\cite{Batatia2025DesignSpace,Batatia2025MACEFoundation}: they encode transferable priors over potential-energy surfaces and can be reused across many systems with limited task-specific retraining. During deployment across the open materials space, uMLIPs can encounter structure-dependent reliability gaps that averaged benchmark metrics often fail to reveal. Atomistic simulation operates in a continuous physical space where elemental combinations, local coordination environments and thermodynamic states vary widely. A model trained on a relatively limited high-fidelity dataset can perform well on a benchmark yet fail on rare or out-of-distribution local environments. For interatomic potentials, such local failures are especially consequential because unreliable force predictions can destabilize molecular dynamics trajectories and lead to temperature drift, atomic collisions or structural collapse\cite{Deng2025Softening,Cui2025TestTimeAdaptation,Riebesell2025MatbenchDiscovery}.

This reliability problem is tightly coupled to the scarcity of high-fidelity training data. Current materials-domain uMLIPs still draw heavily on low- to medium-fidelity labels such as PBE or mixed PBE/PBE+U\cite{Jain2013MaterialsProject,Kirklin2015OQMD,Schmidt2024Alexandria}. r$^2$SCAN data provide higher consistency and accuracy for many solid-state systems, but remain limited in scale, coverage and availability\cite{Furness2020R2SCAN,Kingsbury2022R2SCAN,Huang2025MPR2SCAN,Kuner2025MPALOE}. Directly recomputing r$^2$SCAN labels for tens or hundreds of millions of candidate structures is computationally prohibitive\cite{Kulichenko2024DataGeneration}. Recent datasets have improved this situation from complementary directions. MatPES provides a broad potential-energy-surface dataset constructed with consistent high-fidelity electronic-structure settings\cite{Kaplan2025MatPES}. MP-ALOE expands r$^2$SCAN-level training data with many off-equilibrium and high-force configurations, improving coverage of difficult local environments for uMLIP training\cite{Kuner2025MPALOE}. MAD and MAD-1.5 pursue a different strategy by constructing compact but diverse, high-information atomic datasets that emphasize broad distributional coverage rather than sheer data volume\cite{Mazitov2025MAD,Malosso2026MAD15}. Active learning and high-information-content sampling increase the value of each new DFT calculation\cite{Zhang2019DPGEN,Ko2025DataEfficient,Kim2025Multifidelity}, while multi-fidelity and cross-functional transfer exploit low-fidelity data to improve high-fidelity models\cite{Huang2025CrossFunctional,Kim2025Multifidelity}. Unified data pipelines such as LeMaterial further reduce the cost of organizing large structure pools\cite{Ramlaoui2025LeMatTraj}. These advances improve data quality, data efficiency or data organization, but they do not directly solve a practical data-construction question: given an unlabeled structure from a heterogeneous candidate pool, can a trustworthy r$^2$SCAN-level pseudo-label be generated for it?

The emergence of r$^2$SCAN-aligned pretrained uMLIPs\cite{Huang2025MPR2SCAN,Kuner2025MPALOE}, including MACE\cite{Batatia2025MACEFoundation}, SevenNet-Omni\cite{Kim2026SevenNetOmni}, PET-MAD\cite{Mazitov2025PETMAD} and CHGNet-r$^2$SCAN\cite{Deng2023CHGNet,Huang2025CrossFunctional}, creates a cost-effective route to this question. In this work, these callable universal potentials are referred to as teacher models. They can generate large numbers of candidate energy and force labels without full-space r$^2$SCAN recomputation, but their value depends on structure-wise reliability rather than a global leaderboard ranking. Different teachers are trained with different data sources, target functionals, architectures and optimization strategies, leading to complementary but non-uniform error modes. One teacher may be reliable in a particular chemical or geometric region but fail in another. The key decision is therefore not which teacher is best on average, nor whether teacher predictions should simply be averaged, but which teacher, if any, should be trusted for the current structure.

Related ideas have been explored in machine learning and atomistic modeling, including selective prediction, learning to defer, multi-teacher distillation, mixture-of-experts routing, active-learning uncertainty, committee models and internal confidence estimation\cite{Zhang2019DPGEN,Ko2025DataEfficient,Xu2026EvidentialIP}. These studies show that model uncertainty, disagreement and specialization can improve decision reliability. However, large-scale high-fidelity pseudo-label construction poses a different problem. The goal is not only to train a better model, aggregate several predictions, or decide whether another DFT calculation should be triggered. It is to generate high-confidence labels structure by structure in a large candidate pool while rejecting samples whose risk cannot be controlled. For interatomic potentials, this is also more demanding than ordinary classification routing because the output is a coupled energy--force prediction constrained by physical symmetries, and local force errors have dynamical consequences. This motivates a central question: can the complementarity and disagreement among multiple pretrained uMLIPs be converted into a calibrated accept--reject decision mechanism for high-fidelity materials data construction?

Here we propose Adaptive Multi-Teacher Routing (ATR), a structure-wise routing framework for r$^2$SCAN-level pseudo-label construction. ATR uses a small set of real r$^2$SCAN labels to calibrate multiple pretrained teachers, combines structural descriptors, teacher identity and inter-teacher disagreement, and estimates the reliability of each structure--teacher pair. For each structure, ATR accepts a high-confidence teacher prediction or rejects the structure when no teacher is sufficiently reliable. We deploy ATR on a representative subset sampled from a $10^8$-scale candidate structure space and distil 2.89 million traceable r$^2$SCAN-level pseudo-labels using real r$^2$SCAN labels for only 0.2\% of the deployment candidate structures. A lightweight CHGNet pretrained on the ATR-generated dataset improves held-out r$^2$SCAN prediction and MP-r$^2$SCAN benchmark performance compared with baseline and non-routed controls. Finite-temperature molecular dynamics further shows that ATR pretraining improves dynamical robustness across multiple materials systems, maintaining stable trajectories where baseline simulations undergo catastrophic structural collapse. Together, these results show that ATR converts teacher complementarity and disagreement into calibrated teacher selection, active rejection and high-fidelity materials data expansion, providing a practical route for enlarging the reliable r$^2$SCAN-level training space of uMLIPs.

\section{Results}

\subsection{Framework and dataset overview}

The Adaptive Multi-Teacher Routing (ATR) framework is designed to transform predictions from multiple r\textsuperscript{2}SCAN-aligned pretrained uMLIP teachers into a deployable and traceable high-confidence pseudo-label construction system with an explicit rejection mechanism. The overall workflow consists of five consecutive modules. First, a large-scale candidate structure pool is assembled from public materials databases through a unified data pipeline, while structural sources, trajectory relationships and metadata are preserved. Second, a small number of representative structures are sampled from the candidate pool for real r\textsuperscript{2}SCAN DFT single-point calculations, forming a calibration set for evaluating teacher-model reliability on the target structural distribution. Third, ATR uses this calibration set to learn structure-wise teacher-selection and rejection rules, determining which teacher prediction should be accepted for a given structure or whether the structure should be actively rejected. Fourth, the trained ATR router is deployed to the $10^8$-scale candidate structure pool to generate high-confidence r\textsuperscript{2}SCAN-level pseudo-labels in batches. Finally, the accepted pseudo-label records are used to pretrain a lightweight student model, whose practical benefits are evaluated on a held-out r\textsuperscript{2}SCAN DFT test set, public benchmarks and finite-temperature molecular dynamics simulations. The detailed router training, calibration and deployment procedure is presented separately in \figref{fig:atr-workflow}.

\begin{figure}[!htbp]
\centering
\includegraphics[width=0.98\linewidth,keepaspectratio]{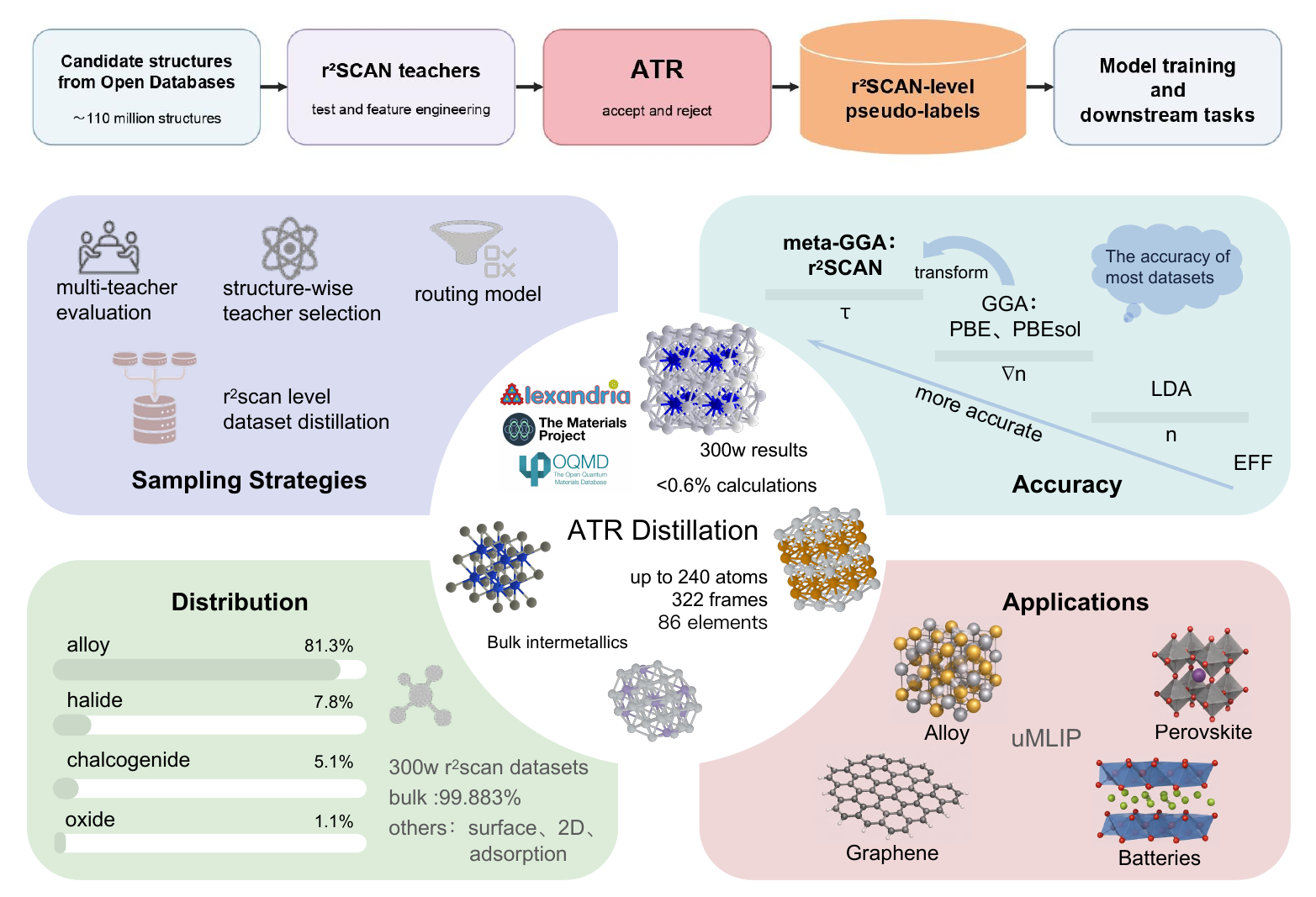}
\caption{Overview of framework and the ATR-distilled r\textsuperscript{2}SCAN-level pseudo-label dataset, including its public data sources, reliability-aware sampling strategy, target-fidelity hierarchy, chemical and structural coverage, dataset scale, and envisioned uMLIP application areas.}
\label{fig:overview}
\end{figure}

\figref{fig:overview} summarizes the resulting data resource, including its public source databases, reliability-aware sampling strategy, target fidelity, chemical and structural coverage, and intended applications. The accepted subset contains nearly 3 million r\textsuperscript{2}SCAN-level pseudo-label records, spans 86 elements, includes structures with up to 240 atoms and 322 trajectory frames. It is dominated by bulk structures and covers major chemical classes including alloys, halides, chalcogenides and oxides, supporting uMLIP pretraining and applications across diverse materials systems.

The key to this workflow is to transform pseudo-label generation for unknown structures from single-teacher inference into structure-wise reliability assessment. In practical data construction, the central question is whether the prediction of a given teacher for the current structure should be accepted. The following results show, in sequence, how the calibration set covers the target materials space, why different teacher models exhibit complementary but non-uniform error modes, how ATR improves pseudo-label quality through routing and active rejection, and how the resulting pseudo-label dataset affects student-model accuracy and dynamical stability.

\FloatBarrier

\subsection{Candidate pool construction and r\textsuperscript{2}SCAN calibration set}

The supervisory signal for ATR comes from a small but highly diverse r\textsuperscript{2}SCAN calibration set. In this work, approximately $1.1\times10^{8}$ candidate structures are first assembled from public sources including the Materials Project, Alexandria, MPTrj and OQMD, and organized into data shards that can be processed in parallel according to source and trajectory information\cite{Jain2013MaterialsProject,Schmidt2024Alexandria,Kirklin2015OQMD,Ramlaoui2025LeMatTraj}. This candidate pool contains PBE-level structures, existing r\textsuperscript{2}SCAN-related structures, and nonequilibrium or trajectory configurations from different database sources. At this stage, the original structures, source identifiers, trajectory associations and source metadata are retained as much as possible. These records support subsequent teacher evaluation, label tracing and large-scale deployment.

Before ATR training, we first perform a preliminary screening of available pretrained uMLIPs based on existing r\textsuperscript{2}SCAN data to form a candidate teacher pool. This teacher pool covers different model families, including SevenNet\cite{Park2024SevenNet,Kim2026SevenNetOmni}, MACE\cite{Batatia2022MACE,Batatia2025MACEFoundation}, CHGNet\cite{Deng2023CHGNet,Huang2025CrossFunctional} and PET-MAD\cite{Mazitov2025PETMAD}. We then sample approximately $1.8\times10^{4}$ structures from the candidate structure pool and perform real r\textsuperscript{2}SCAN DFT single-point calculations to construct the ATR calibration set. The purpose of this set is not to reproduce a fixed benchmark, but to provide supervised signals that are as close as possible to the target distribution for subsequent deployment on the $10^8$-scale candidate pool. During sampling, repeated contributions from adjacent frames within the same trajectory are preferentially avoided, while broad coverage of elemental compositions, chemical systems and structure types is pursued. Statistical analysis shows that this calibration set is dominated by highly deduplicated single-frame structures: approximately 18,000 structures correspond to 17,792 unique material group IDs, and 98.9\% of material groups contain only one frame. This property reduces optimistic bias in teacher evaluation and routing training caused by near-duplicate configurations.

In terms of compositional and structural distributions, this calibration set shows strong diversity. It covers 85 elements, including main-group elements, transition metals, lanthanides, actinides and heavy elements. The number of unique chemical systems reaches 14,743, and the number of unique reduced formulas reaches 17,522; even the most common chemical system accounts for only 0.29\% of all samples. In terms of compositional complexity, the samples are dominated by ternary and quaternary systems, while also including binary systems and a small number of more complex multicomponent systems. In terms of structural size, the median number of atoms is 11, 90\% of structures contain no more than 20 atoms, and 95\% contain no more than 28 atoms; a small number of larger structures are also retained. Overall, although this calibration set is much smaller than the candidate pool, it provides broad supervisory coverage in elements, chemical systems and structural complexity. This coverage establishes the basis for subsequent structure-wise learning of teacher reliability.

\begin{figure}[!htbp]
\centering
\includegraphics[width=0.95\linewidth,height=0.82\textheight,keepaspectratio]{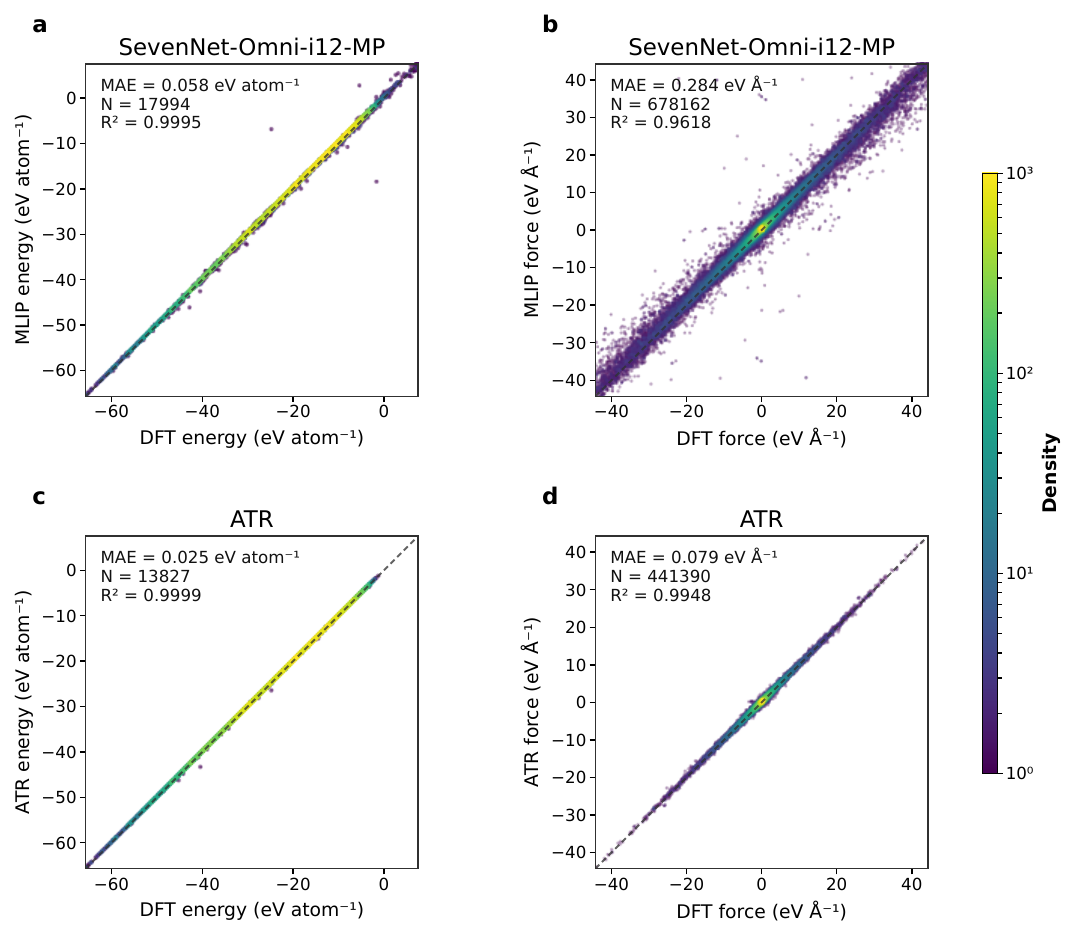}
\caption{\textbf{Strong-teacher and ATR predictions on the r\textsuperscript{2}SCAN calibration set.}
\textbf{a,b,} Comparison of per-atom energies (\textbf{a}) and Cartesian force components (\textbf{b}) between the SevenNet-Omni-i12-MP\cite{Kim2026SevenNetOmni} teacher predictions and r\textsuperscript{2}SCAN DFT reference values.
\textbf{c,d,} Comparison of per-atom energies (\textbf{c}) and Cartesian force components (\textbf{d}) between predictions after ATR routing and ground truth. Color indicates point density, and the dashed lines denote ideal parity. Each panel reports the MAE, the number of evaluated entries $N$, and $R^2$. For the energy panels, $N$ denotes the number of structures; for the force panels, $N$ denotes the number of Cartesian force components. The smaller values of $N$ in the ATR panels reflect active rejection.}
\label{fig:teacher-evaluation}
\end{figure}

On this calibration set, we systematically evaluate the energy and force prediction errors of all ten candidate teacher models. The complete comparison shows that the teachers do not follow a simple global performance ranking: models with similar overall MAEs can exhibit distinct error modes across element combinations, local geometric environments or structural clusters. Some teachers are relatively stable for one class of structures but show systematic bias for another. Among these models, SevenNet-Omni-i12-MP\cite{Kim2026SevenNetOmni} provides a representative strong single-teacher reference on the current target distribution, and its comparison with the predictions retained after ATR routing and active rejection is shown in \figref{fig:teacher-evaluation}. Other teachers still provide complementary information in local regions. These observations indicate that a single average-error ranking is insufficient for large-scale pseudo-label construction and directly motivate the core task of ATR: learning reliable structure-wise selection among multiple candidate teachers and actively rejecting structures when all teachers are unreliable.

\FloatBarrier

\subsection{Structure-wise teacher routing with rejection}

\begin{figure}[!htbp]
\centering
\includegraphics[width=0.95\linewidth,height=0.82\textheight,keepaspectratio]{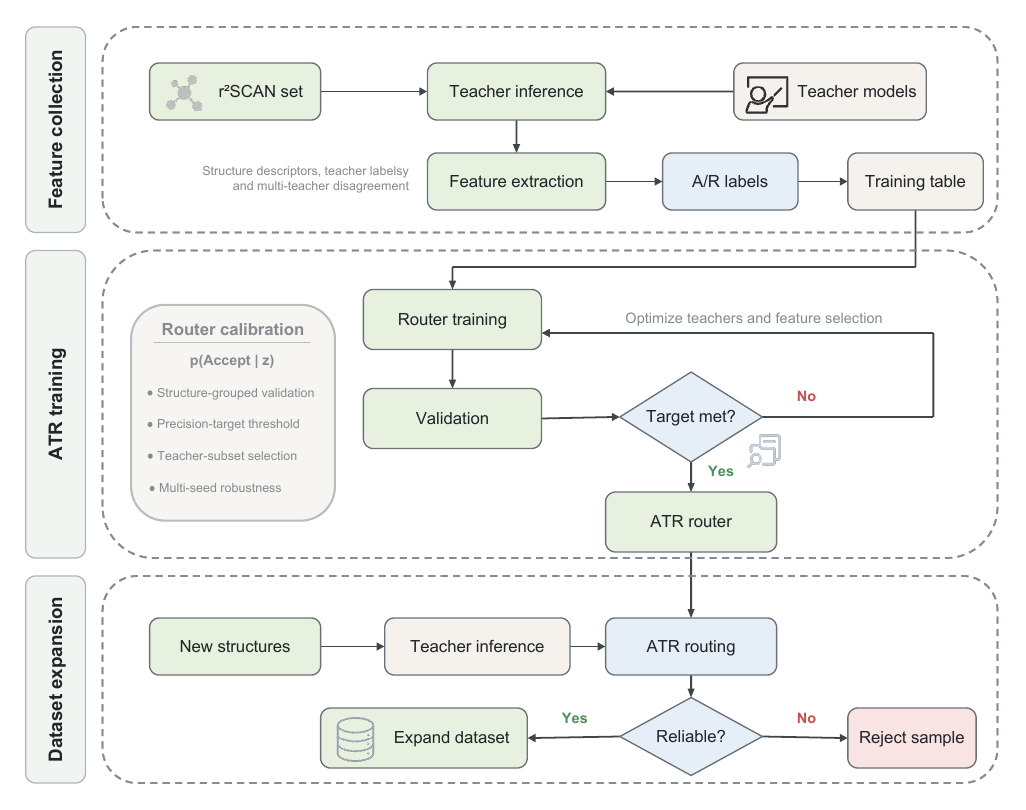}
\caption{\textbf{ATR training, calibration and dataset-expansion workflow.}
The feature-collection stage combines the r\textsuperscript{2}SCAN calibration set, teacher-model inference, structural descriptors, teacher identity and multi-teacher disagreement features to construct structure--teacher Accept/Reject labels and the router training table. During ATR training, structure-grouped validation, acceptance-probability calibration, precision-target thresholding, teacher-subset selection and multi-seed robustness checks are used to select the final router. After training, the router evaluates teacher predictions for new structures, accepts a selected prediction when the reliability criterion is satisfied, and otherwise assigns the structure to the reject pool.}
\label{fig:atr-workflow}
\end{figure}

Based on the multi-teacher error complementarity observed on the calibration set, ATR formulates pseudo-label generation as an acceptability classification problem over structure--teacher pairs. For each candidate structure $x_i$ and teacher model $M_k$, ATR determines whether the energy/force predictions of this teacher satisfy the predefined quality criterion for the current structure. The final deployment in this work adopts a binary Accept/Reject definition: a structure--teacher pair is defined as Accept when it simultaneously satisfies $|\Delta E|<0.1$ eV/atom and $F_{\max}<1.0$ eV/\AA{}; otherwise, it is defined as Reject. For an unlabeled structure, ATR selects the teacher with the highest acceptance probability among all candidate teachers. If the highest acceptance probability is still lower than the calibrated threshold, the structure is assigned to the reject pool rather than being forced to receive a pseudo-label. The corresponding training, calibration and deployment workflow is summarized in \figref{fig:atr-workflow}.

\FloatBarrier

ATR jointly uses structural descriptors, multi-teacher disagreement features, teacher-identity features and risk-control features. In this way, the decision of whether to accept a teacher prediction becomes a learnable structure-wise classification problem. Risk control is particularly important. When multiple strong teachers give consistent predictions on the same structure, that structure is more likely to lie in a high-confidence region, although consistency alone does not guarantee correctness. When energy or force predictions differ substantially among teachers, the structure is more likely to be located in a model blind spot, a locally anomalous configuration or an out-of-distribution region. Therefore, ATR does not make an independent decision based on the internal confidence of a single teacher. Instead, it learns structure-wise reliability from the relative behavior of multiple teachers on the same structure.

We compare several classifiers as lightweight tree-model backends, including RandomForest, XGBoost and LightGBM\cite{Chen2016XGBoost,Ke2017LightGBM} (\tabref{tab:backend}). All three implement similar structure-wise screening functions and yield similar trends in coverage, precision, energy MAE and force MAE. This result indicates that the main benefit of ATR comes from the problem definition, feature construction and active-rejection mechanism, rather than from a specific tree model. Among these backends, XGBoost provides a stable overall balance and is therefore selected as the classifier for the subsequent ATR implementation.

\begin{table}[H]
\centering
\caption{Comparison of lightweight router backends under the same ATR feature set and evaluation protocol.}
\label{tab:backend}
\begin{tabular}{lcccc}
\toprule
Backend & Coverage & Precision & Energy MAE (eV/atom) & Force MAE (eV/\AA) \\
\midrule
RandomForest & 0.5285 & 0.9038 & 0.0234 & 0.1105 \\
XGBoost & 0.5456 & 0.9042 & 0.0224 & 0.1048 \\
LightGBM & 0.5267 & 0.9046 & 0.0227 & 0.0996 \\
\bottomrule
\end{tabular}
\end{table}

We next examine the effect of teacher-set size on routing quality (\tabref{tab:teacher-subsets}). The results show that more teachers do not necessarily lead to better pseudo-labels. The top1 configuration has higher coverage but lower precision and more anomalous structures. The top3 configuration alleviates the lack of complementarity in a single-teacher setting. Although the top8 configuration uses the largest number of candidate teachers and expands the input information, additional weak teachers may introduce noise and contaminate the disagreement features, leading to degraded performance. In contrast, top5 achieves a better overall balance under the current target distribution and teacher-pool conditions. Its coverage is 77.44\%, precision is 92.81\%, force MAE decreases to 0.08168 eV/\AA{}, and the number of anomalous structures is reduced to the minimum. This result shows that the benefit of ATR does not come from blindly stacking more teachers, but from balancing teacher strength, complementarity and noise control.

\begin{table}[H]
\centering
\caption{Screening performance of ATR with different numbers of top teacher models.}
\label{tab:teacher-subsets}
\resizebox{\textwidth}{!}{%
\begin{tabular}{lcccccc}
\toprule
Teacher Subset & Coverage & Precision & Accepted Structures & Energy MAE (eV/atom) & Force MAE (eV/\AA) & Anomalous Structures \\
\midrule
top1 & 79.49\% & 90.22\% & 14,309 & 0.02985 & 0.09072 & 49 \\
top3 & 79.69\% & 91.42\% & 14,344 & 0.02682& 0.08741 & 22 \\
top5 & 77.44\% & \textbf{92.81\%} & 13,939 & \textbf{0.02681}& \textbf{0.08168} & 3\\
top8 & 79.71\% & 90.93\% & 14,348 & 0.02817 & 0.08764 & 20 \\
\bottomrule
\end{tabular}%
}
\end{table}

\begin{figure}[!htbp]
\centering
\includegraphics[width=0.92\linewidth,height=0.74\textheight,keepaspectratio]{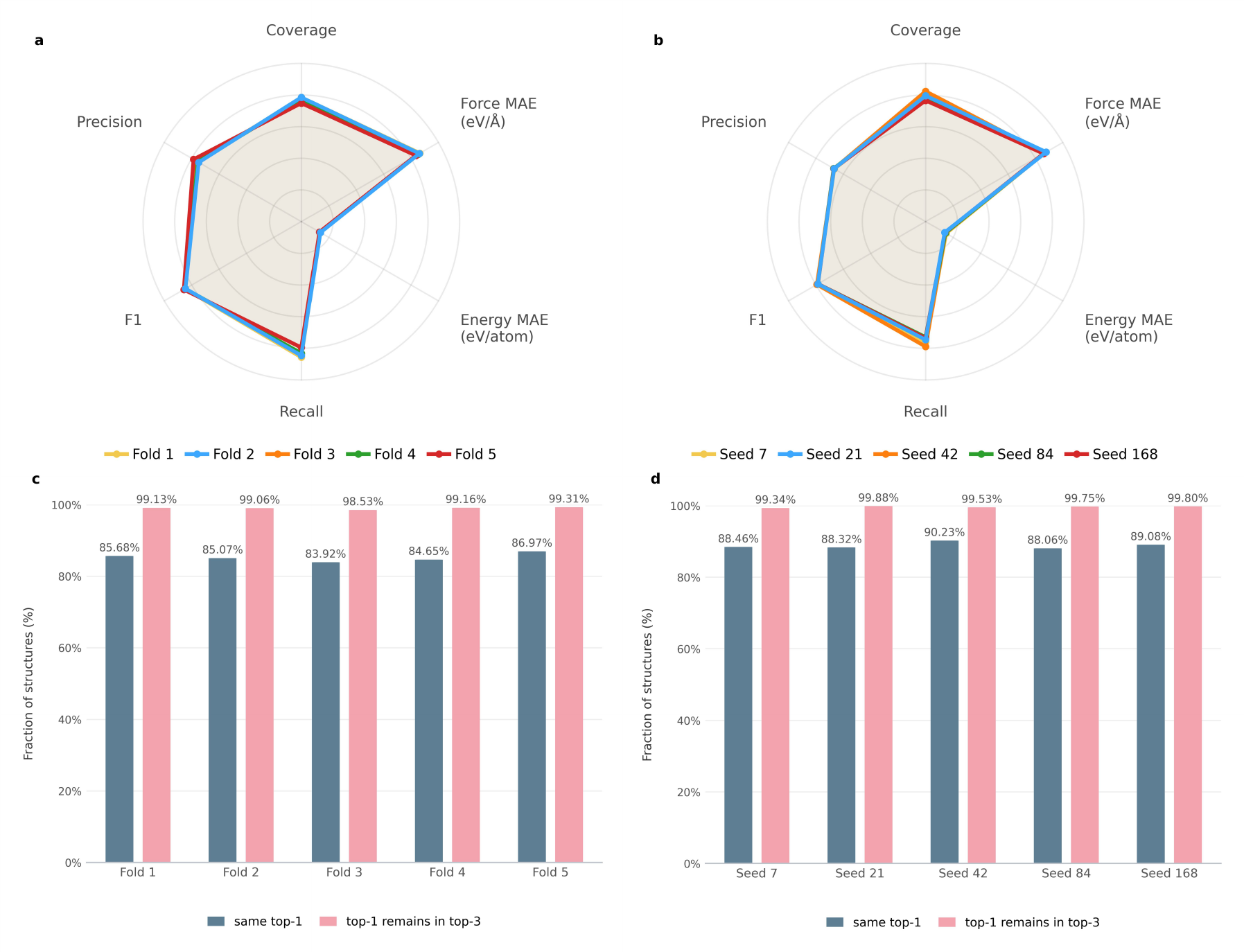}
\caption{\textbf{Robustness of ATR routing across data splits and random seeds.}
\textbf{a,b,} Router-level performance under grouped five-fold cross-validation (\textbf{a}) and five random seeds (\textbf{b}), summarized by coverage, precision, recall, F1, energy MAE and force MAE.
\textbf{c,d,} Corresponding teacher-selection stability across the cross-validation folds (\textbf{c}) and random seeds (\textbf{d}). Dark bars show the fraction of structures assigned the same top-1 teacher as the reference result, whereas light bars show the fraction for which the reference top-1 teacher remains within the top-3 candidates.}
\label{fig:atr-performance}
\end{figure}

We further evaluate the stability of ATR through grouped five-fold cross-validation and multi-random-seed experiments (\figref{fig:atr-performance}). In the five-fold cross-validation, the distributions of coverage, precision, F1, energy MAE and force MAE are similar across folds, indicating that ATR performance does not depend on a fortuitous data split. Teacher selection is also highly consistent. In the five-fold experiments, the fraction of structures with the exact same top-1 teacher is approximately 83.9\%--87.0\%, and the fraction for which the reference top-1 teacher remains within the top-3 candidates reaches 98.5\%--99.3\%. In the multi-random-seed experiments, these two fractions are approximately 88.1\%--90.2\% and 99.3\%--99.9\%, respectively. Thus, even when ATR does not always output exactly the same unique teacher for every structure, it almost always routes structures to the same group of high-quality candidate teachers. This stability supports reproducibility during large-scale deployment.
\FloatBarrier
Finally, this work adopts the ATR configuration with top5 teachers plus lightweight postprocessing as the practical deployment strategy. This strategy achieves a conservative but reliable balance between acceptance rate and pseudo-label purity, with a coverage of 0.7744, a precision of 0.9281, and energy and force MAEs of 0.0268 eV/atom and 0.0817 eV/\AA{}, respectively. Overall, by explicitly modeling structure-wise reliability through a strong-teacher set, structure-wise routing and active rejection, ATR converts multi-teacher complementarity into a deployable capability for generating r\textsuperscript{2}SCAN-level pseudo-labels.

\subsection{Large-scale deployment, distributional coverage and student-model validation}

After ATR calibration and final configuration selection are completed, we deploy ATR to the large-scale candidate structure pool to test whether this structure-wise routing strategy can generalize from small-scale real r\textsuperscript{2}SCAN supervision to the unlabeled open materials space. To facilitate subsequent validation through student-model training, we sample $7,674,241$ structures from the full PBE candidate data as a representative subset and run the final ATR workflow. This produces $2,893,331$ high-confidence accepted samples as the pseudo-label subset. The acceptance rates differ substantially across shards of the subset: high-acceptance shards exceed 90\%, whereas low-acceptance shards are approximately 18\%, with an overall acceptance level of approximately 40\%. This result shows that ATR does not retain structures randomly at a fixed ratio. Instead, it adjusts acceptance behavior according to structural source, local configurational difficulty and the level of teacher agreement, thereby performing structure-wise risk control in real deployment.

This accepted subset is then divided into training and validation sets, while teacher sources, original structural metadata, routing decisions and source identifiers are retained. To avoid data leakage, both subsampling and data splitting are performed at the trajectory or immutable-id level, so that configurations from the same trajectory or material group are not shared across the training, validation and test splits. Among the 2.89 million accepted samples, approximately 2.6 million structures are used for student-model pretraining after the train/validation split and quality-control filtering. The resulting million-scale pseudo-label subset is therefore accompanied by records of teacher selection, routing decisions and source metadata, providing traceable r\textsuperscript{2}SCAN-level training data.

\begin{figure}[!htbp]
\centering
\includegraphics[width=0.95\linewidth,height=0.82\textheight,keepaspectratio]{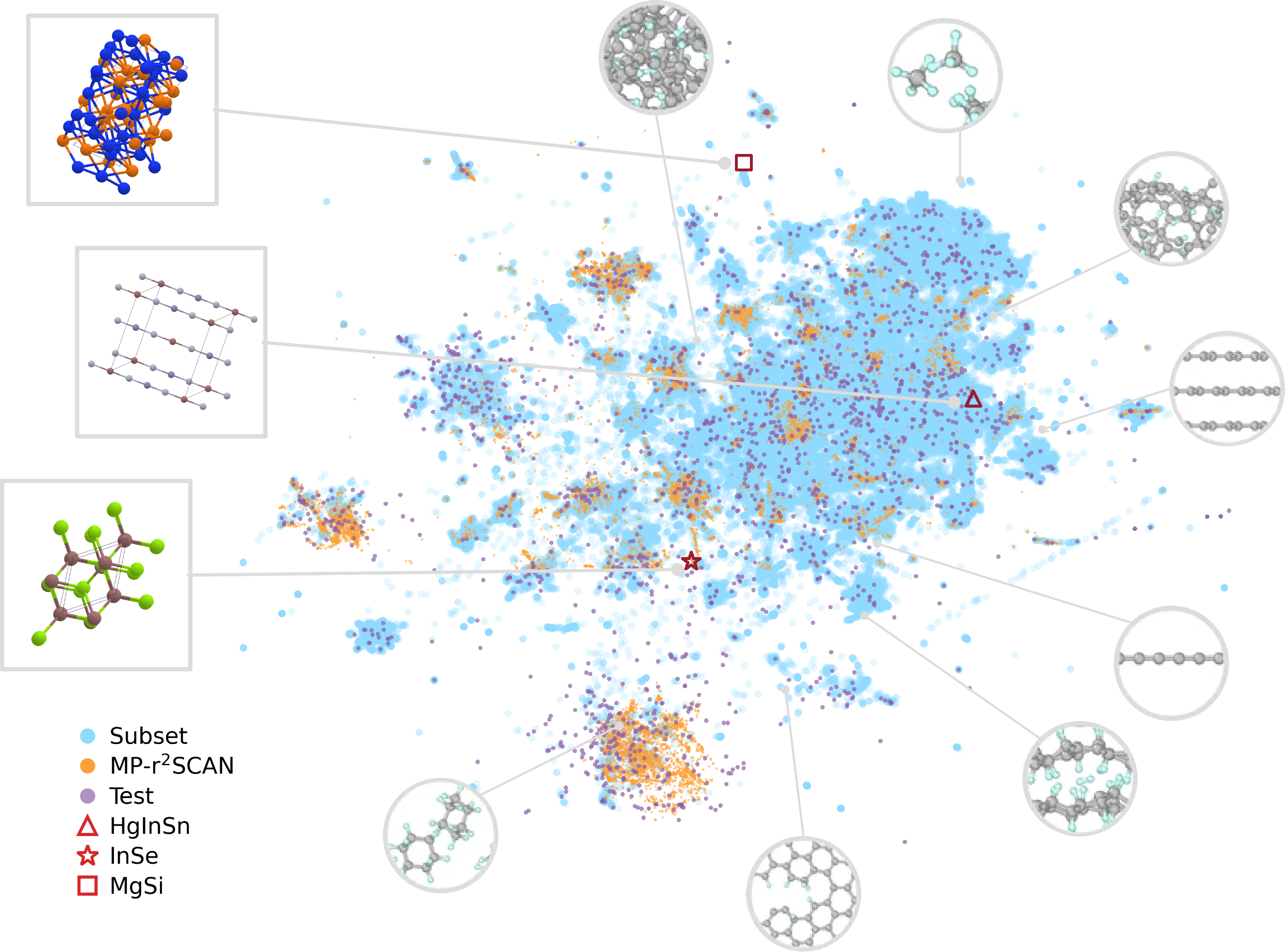}
\caption{\textbf{Material structure embedding of the ATR pseudo-label subset and evaluation data.}
UMAP projection of the ATR-selected high-confidence pseudo-label subset (light blue), MP-r\textsuperscript{2}SCAN structures (orange), and all 2,000 held-out r\textsuperscript{2}SCAN DFT test structures (purple). Red markers identify the HgInSn, InSe and MgSi systems used for MD validation. Because different sampling ratios are used for the plotted datasets, the visualization is intended to compare relative distributional positions rather than absolute probability densities in the original feature space.}
\label{fig:dataset-umap}
\end{figure}

To compare the distributional relationships among the ATR-selected high-confidence subset, existing r\textsuperscript{2}SCAN data and the evaluation structures, we construct a chemical--structural UMAP visualization (\figref{fig:dataset-umap})\cite{McInnes2018UMAP}. The Full candidate dataset defines the global materials-space reference used for the embedding. The figure instead shows coverage-preserving 10\% samples of the ATR-selected Subset and MP-r\textsuperscript{2}SCAN, together with all 2,000 held-out r\textsuperscript{2}SCAN DFT test structures. The three MD validation systems HgInSn, InSe and MgSi are marked with special symbols. Because the plotted datasets are sampled differently, the visualization is intended to compare relative distributional positions rather than absolute point densities. The ATR subset overlaps the existing MP-r\textsuperscript{2}SCAN distribution while also occupying additional regions that contain held-out test structures, enabling evaluation of student-model generalization across a broader target distribution.

\FloatBarrier

To verify that ATR pseudo-labels have measurable training value, rather than merely favorable screening metrics, we perform pretraining and downstream fine-tuning experiments using CHGNet as a lightweight student model (\tabref{tab:student-validation})\cite{Deng2023CHGNet}. The results show that ATR pseudo-label pretraining substantially improves the generalization ability of the student model relative to the baseline. The without ATR 1 and 2 controls further show that simply increasing the number of pretraining structures or changing the training protocol is insufficient to reproduce the benefit of ATR. Although the two without ATR variants improve over the baseline to some extent, both are worse than the ATR student model on the held-out DFT test set and the MP-r\textsuperscript{2}SCAN benchmark\cite{Huang2025MPR2SCAN,Riebesell2025MatbenchDiscovery,Kuner2025MPALOE}. This indicates that the performance improvement mainly comes from ATR-calibrated high-confidence r\textsuperscript{2}SCAN-level pseudo-labels, rather than from a simple increase in data volume or changes to the training protocol.

\begin{table}[H]
\centering
\caption{Performance comparison of student models and teacher references on held-out r\textsuperscript{2}SCAN DFT structures and the MP-r\textsuperscript{2}SCAN benchmark.}
\label{tab:student-validation}
\resizebox{\textwidth}{!}{%
\begin{tabular}{lcccc}
\toprule
Model & Held-out DFT Energy MAE & Held-out DFT Force MAE & MP-r\textsuperscript{2}SCAN Energy MAE & MP-r\textsuperscript{2}SCAN Force MAE \\
\midrule
baseline& 1.634 & 2.847 & 0.0788 & 0.0661 \\
without ATR1& 1.027& 1.895& 0.0294& 0.0242\\
without ATR2& 1.375& 1.628& 0.0273&0.0240\\
student& \textbf{0.074}& \textbf{0.069}& \textbf{0.0148}& \textbf{0.0216}\\
CHGNet-r\textsuperscript{2}SCAN\cite{Huang2025CrossFunctional}& 0.455 & 2.053 & 0.017 (0.0159) &0.038 (0.0211) \\
7net-omni-i12-mp\_r\textsuperscript{2}SCAN\cite{Kim2026SevenNetOmni} & 0.058 & 0.307 & 0.0068 & 0.0136 \\
7net-omni-i12-matpes\_r\textsuperscript{2}SCAN\cite{Kim2026SevenNetOmni} & 0.446 & 0.311 & 0.4326 & 0.0168 \\
\bottomrule
\end{tabular}%
}

\vspace{2pt}
\noindent
\begin{minipage}{\textwidth}
\setlength{\parindent}{0pt}
\fontsize{6.5pt}{7.4pt}\selectfont
Energy MAE is reported in eV/atom, and force MAE is reported in eV/\AA{}.
The baseline is a CHGNet model trained only on MP-r\textsuperscript{2}SCAN without ATR pseudo-label pretraining.
The without-ATR controls are non-routed pretraining baselines on the same 2.6M structure pool, including supervised pretraining and self-/supervised pretraining followed by MP-r\textsuperscript{2}SCAN fine-tuning.
The student model is pretrained on ATR-selected r\textsuperscript{2}SCAN-level pseudo-labels and then fine-tuned on MP-r\textsuperscript{2}SCAN.
CHGNet-r\textsuperscript{2}SCAN is the official CHGNet-r\textsuperscript{2}SCAN checkpoint, and the SevenNet entries are SevenNet-Omni-i12 teacher checkpoints aligned with MP-r\textsuperscript{2}SCAN or MatPES-r\textsuperscript{2}SCAN data; these teacher checkpoints are included as reference models rather than downstream baselines.
\end{minipage}
\end{table}

The comparison with strong teacher models further shows that ATR high-confidence data do not merely copy the global behavior of a single teacher. Instead, they can yield a student model whose selected metrics match or approach those of strong teacher references. On the held-out test set, the energy accuracy of the student model is of the same order as that of the strongest teacher, while its force prediction is better than those of multiple teachers. On the MP-r\textsuperscript{2}SCAN benchmark, the complete training splits of public checkpoints are not fully traceable, and the test set used in this work may overlap with the training data of the teacher models. Nevertheless, the results show that the student can approach, and on some metrics exceed, the teacher models. Overall, the pseudo-labels generated by ATR are not simply products of large-scale model inference. They can serve as transferable high-accuracy priors and substantially improve the generalization ability of a lightweight student model on r\textsuperscript{2}SCAN-level prediction tasks.

\subsection{MD simulation validation}

\begin{figure}[!htbp]
\centering
\includegraphics[width=1.0\linewidth]{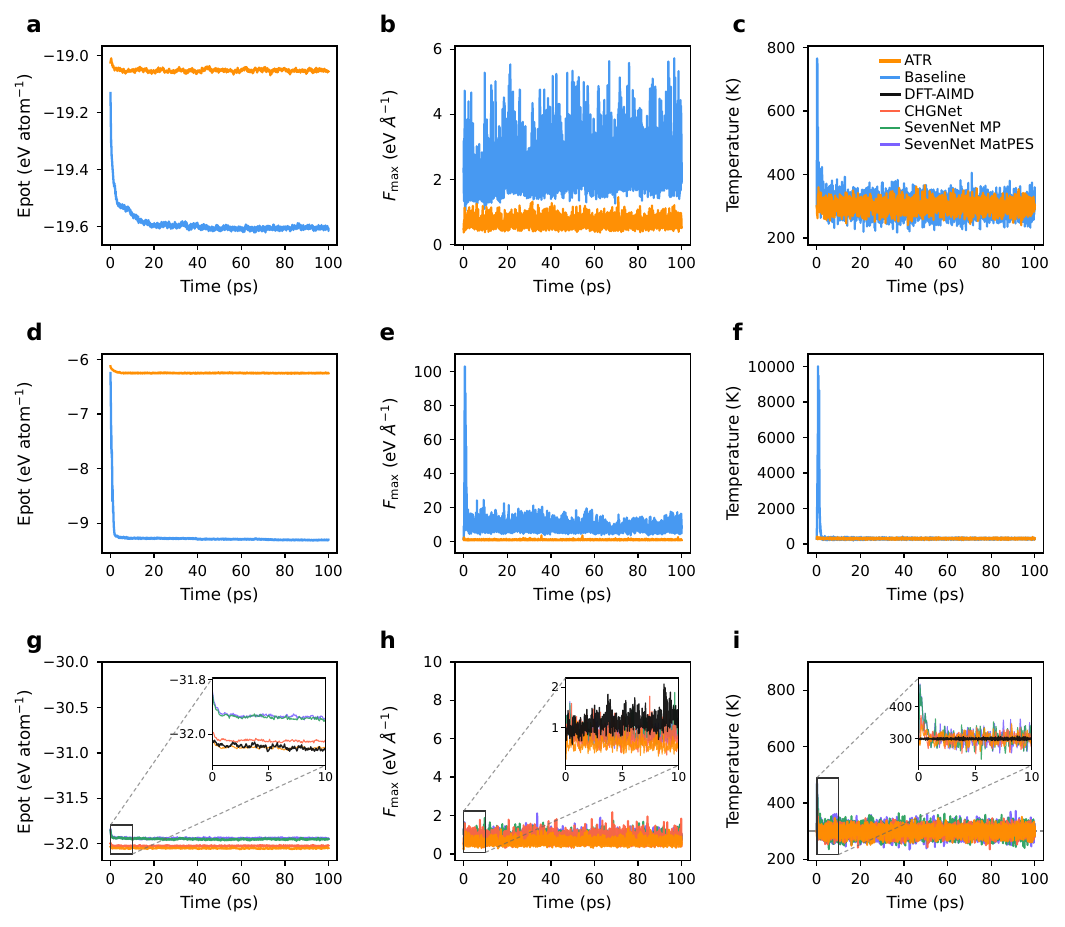}
\caption{\textbf{Molecular dynamics validation of the ATR student model.}
\textbf{a--c,} 300 K MLMD trajectories for InSe comparing the ATR student model and the CHGNet baseline: potential energy per atom (\textbf{a}), maximum atomic force in each frame (\textbf{b}), and temperature (\textbf{c}).
\textbf{d--f,} Corresponding 300 K MLMD results for MgSi: potential energy per atom (\textbf{d}), maximum atomic force in each frame (\textbf{e}), and temperature (\textbf{f}).
\textbf{g--i,} HgInSn trajectories comparing the ATR student model, the baseline, CHGNet-r\textsuperscript{2}SCAN\cite{Huang2025CrossFunctional}, SevenNet-Omni-i12-MP, SevenNet-Omni-i12-MatPES\cite{Kim2026SevenNetOmni} and the r\textsuperscript{2}SCAN DFT-AIMD reference: potential energy per atom (\textbf{g}), maximum atomic force in each frame (\textbf{h}), and temperature (\textbf{i}). The main panels show the 100 ps MLMD trajectories, whereas the insets enlarge the first 10 ps to facilitate comparison with the r\textsuperscript{2}SCAN DFT-AIMD reference. The displayed temperature ranges are truncated where necessary for visual clarity; complete extrema are reported in Table~\ref{tab:mlmd-stability}.}
\label{fig:mlmd-diagnostics}
\end{figure}

Static-structure test errors measure a model's ability to predict single-point energies and forces, but a materials potential must ultimately remain stable in dynamical simulations. To this end, we select three 300 K r\textsuperscript{2}SCAN AIMD reference systems, HgInSn, MgSi and InSe, and perform MLMD validation for the ATR student model and the corresponding baseline (\figref{fig:mlmd-diagnostics}, \tabref{tab:mlmd-stability}). In \figref{fig:mlmd-diagnostics}, panels \textbf{a--c} and \textbf{d--f} show the InSe and MgSi trajectories, respectively, whereas panels \textbf{g--i} compare the HgInSn trajectories with the teacher references and the r\textsuperscript{2}SCAN DFT-AIMD reference. The main panels show 100 ps MLMD trajectories, and the insets enlarge the first 10 ps. The three systems are complementary in their chemical bonding and data-coverage relationships. HgInSn is a ternary soft-metal alloy located in a region well covered by the ATR subset but relatively sparse in MP-r\textsuperscript{2}SCAN. InSe is closer to the existing MP-r\textsuperscript{2}SCAN coverage region, whereas MgSi lies near a lower-density region with stronger extrapolative character. This set of tests therefore probes student-model dynamics in an extended-coverage region, an existing high-accuracy coverage region and a low-coverage region.

\FloatBarrier

The MLMD results show that the student model maintains stable 300 K trajectories in all three systems, whereas the baseline shows clear instability in HgInSn and MgSi. In MgSi, the maximum temperature of the baseline rises to 44,724 K, the maximum atomic force reaches 94.11 eV/\AA{}, and the minimum interatomic distance decreases to 0.20 \AA{}, indicating catastrophic structural collapse. In contrast, the MD results of the student model remain physically reasonable. In InSe, the instability of the baseline is weaker, but the ATR student model still shows a lower maximum force, a temperature distribution closer to the target temperature and more stable nearest-neighbor distances. In HgInSn, the ATR student model also maintains a stable trajectory close to 300 K, whereas the baseline exhibits temperature spikes, force anomalies and reduced nearest-neighbor distances. For the HgInSn system, we further compare against different teacher models on a 10 ps r\textsuperscript{2}SCAN DFT-AIMD reference trajectory\cite{Furness2020R2SCAN,Kaplan2025MatPES}. The energy MAE of the student model is 17.9 meV/atom, substantially lower than those of the other evaluated models. Together with the MLMD stability metrics, this result shows that the student model maintains stable temperature, force and nearest-neighbor distance in HgInSn and obtains an energy curve closer to r\textsuperscript{2}SCAN DFT-AIMD in this complex soft-metal alloy system.

\begin{table}[H]
\centering
\caption{MD stability comparison of ATR, baseline and teacher-reference models.}
\label{tab:mlmd-stability}
\resizebox{\textwidth}{!}{%
\begin{tabular}{lccccc}
\toprule
System & Model & $T_{\mathrm{mean}}/T_{\mathrm{max}}$ (K) & $F_{\max,\mathrm{mean}}/F_{\max,\mathrm{max}}$ (eV/\AA) & min $d_{\min}$ (\AA) & Energy MAE \\
\midrule
MgSi & student& \textbf{308 / 422} & \textbf{1.04 / 2.43} & 1.90 &-\\
MgSi & baseline& 1,045 / 44,724 & 12.20 / 94.11 & 0.20 &-\\
InSe & student& \textbf{302 / 360} & \textbf{0.71 / 1.35} & 2.20 &-\\
InSe & baseline& 331 / 767 & 2.14 / 6.76 & 1.98 &-\\
HgInSn & student& \textbf{300.10 / 344}& \textbf{0.62 / 1.34}& 2.24 &\textbf{17.9} \\
HgInSn & baseline& 306.28 / 1,207& 3.23 / 6.98& 1.91 &174.8 \\
HgInSn & SevenNet-Omni-i12-MP\cite{Kim2026SevenNetOmni} & 300.73 / 467.5& 0.86 / 1.87& 2.56&193.7 \\
HgInSn & SevenNet-Omni-i12-MatPES\cite{Kim2026SevenNetOmni} & 300.76 / 470.8& 0.87 / 2.11& 2.55&202.0 \\
HgInSn & CHGNet-r\textsuperscript{2}SCAN\cite{Huang2025CrossFunctional} & 300.19 / 372.3& 0.78 / 2.17& 2.43&43.2 \\
\bottomrule
\end{tabular}%
}

\vspace{2pt}
\noindent
\begin{minipage}{\textwidth}
\setlength{\parindent}{0pt}
\fontsize{6.5pt}{7.4pt}\selectfont
$T$ denotes temperature, $F_{\max}$ denotes the maximum atomic force in each frame, and $d_{\min}$ denotes the minimum interatomic distance along the trajectory.
MD stability statistics are computed from 100 ps trajectories.
Energy MAE is reported in meV/atom and is calculated as the MAE between the model-predicted and DFT energies at each time point.
\end{minipage}
\end{table}

These MD validation results show that the value of ATR extends beyond static screening metrics and held-out single-point MAEs to the dynamical stability of the potential-energy surface. The instability of the baseline in MgSi and HgInSn indicates that a lightweight student model trained only on MP-r\textsuperscript{2}SCAN data cannot easily cover certain local regions of the potential-energy surface. By expanding the r\textsuperscript{2}SCAN-level local environments visible to the student model through high-confidence pseudo-labels, ATR enables more stable dynamical behavior under different coverage conditions. Overall, the student-model validation and MD results jointly demonstrate that ATR pseudo-label data can transfer improvements in static accuracy to dynamical robustness in practical atomistic simulations.

\FloatBarrier
\section{Discussion}

The central contribution of ATR is to shift high-accuracy materials data construction from a model-centered strategy to a decision-centered one. In pretrained-uMLIP-assisted data construction or downstream system modeling, common strategies include selecting a model with strong average benchmark performance to distil or annotate a pretraining dataset\cite{Riebesell2025MatbenchDiscovery,Fu2025eSEN,Wood2025UMA,Rhodes2025OrbV3}, fine-tuning a pretrained potential on a target system\cite{Cui2025TestTimeAdaptation,Xu2026EvidentialIP}, or using committee uncertainty to decide when additional DFT calculations should be added\cite{Zhang2019DPGEN,Ko2025DataEfficient}. Model generalization is then often assessed by downstream average errors\cite{Focassio2024Surfaces,Deng2025Softening}. These practices are useful, but average errors do not determine whether a pseudo-label is reliable for a specific structure, and multiple models may fail in the same local region of the potential-energy surface. ATR addresses this data-construction risk by using a small number of real r$^2$SCAN annotations to learn the acceptability of each structure--teacher pair. The methodological focus is the calibrated acceptability formulation: which teacher prediction should be accepted for a given structure and when the structure should be rejected.

This design differs from simple model ensembling, voting or prediction averaging. ATR evaluates local reliability using structural features, teacher identity and inter-teacher disagreement, rather than smoothing outputs across teachers. The teacher-subset experiments clarify why this distinction matters. More teachers do not necessarily yield better pseudo-labels: too few teachers limit complementarity, whereas too many weak teachers can introduce noisy disagreement features and reduce the router's ability to distinguish reliable samples from risky ones. Under the present target distribution, the top-5 strong-teacher subset provides the best balance between precision, force error and anomaly control. Constructing the teacher set is therefore itself a calibration problem involving model strength, complementarity and noise control. The top-5 configuration should be viewed as an empirically calibrated choice for this deployment setting, not as a universal rule for all uMLIP routing tasks.

From the data-construction perspective, ATR acts as a structured teacher-assisted distillation mechanism rather than a tool for copying any single teacher\cite{Cui2024GeometryPretraining,Batatia2025MACEFoundation,Kim2026SevenNetOmni,Mazitov2025PETMAD}. It identifies locally reliable predictions from different teachers and converts them into trainable r$^2$SCAN-level pseudo-labels. Student-model validation and ablation experiments separate the effect of ATR from longer training or larger data volume: non-routed controls using the same large structure pool and comparable pretraining workflows still lag behind the ATR student. This comparison indicates that the improvement comes primarily from calibrated routing and active rejection, which suppress low-confidence pseudo-labels before they are transferred into the student potential. Large-scale deployment shows non-random acceptance behavior across data shards, consistent with structure-wise risk control. The UMAP analysis further shows that the ATR subset overlaps existing MP-r$^2$SCAN data while extending into additional regions. These regions contain held-out test structures and cover the neighborhoods of the MD validation systems, supporting its role in expanding r$^2$SCAN-level training coverage.

The practical importance of this decision-driven data construction becomes most visible in molecular dynamics. For interatomic potentials, a small number of anomalous forces or distorted short-range interactions can destabilize long-time integration even when static MAEs appear acceptable. In the HgInSn, MgSi and InSe tests, ATR pseudo-label pretraining improves not only held-out single-point prediction but also temperature control, force stability and nearest-neighbor-distance behavior in 300 K trajectories. Baseline simulations undergo catastrophic structural collapse in challenging systems, whereas the ATR student maintains stable dynamics. This indicates that active rejection can remove high-risk pseudo-labels before they are transferred into the student potential, expanding the r$^2$SCAN-level local environments visible during training while preserving physical reliability.

The current ATR implementation uses accept/reject criteria based on single-point energies and atomic forces, which matches the objective of r$^2$SCAN-level pseudo-label construction. Future extensions should incorporate additional physical constraints such as stress, magnetism, relaxation stability, trajectory consistency, RDFs, diffusion, elasticity and phase stability, and should validate longer time scales, higher temperatures, defects, surfaces and interfaces. These quantities can be integrated into future label definitions, routing features and risk-control rules, extending ATR from energy--force pseudo-label filtering toward more complete physical reliability control.

In conclusion, we propose ATR as a decision-driven framework for high-fidelity materials data expansion with pretrained uMLIP teachers. ATR connects limited real r$^2$SCAN annotations, existing universal potentials, large unlabeled structure pools and lightweight student-model training into a deployable closed loop. By making the decision of when to accept a teacher prediction and when to reject a structure explicit, ATR provides a practical route for constructing reliable r$^2$SCAN-level training data and improving the reliablity and the generalization of next-generation high-accuracy uMLIPs.

\section{Methods}

\subsection{Candidate structure collection}
Candidate structures are automatically assembled from public databases including the Materials Project, Alexandria, MPTrj and OQMD through a LeMaterial-Fetcher-like pipeline\cite{Jain2013MaterialsProject,Schmidt2024Alexandria,Kirklin2015OQMD,Ramlaoui2025LeMatTraj}. This workflow addresses inconsistencies in format, metadata and accessibility across multi-source materials trajectory data. It unifies structural objects, trajectory relationships, source identifiers, calculation levels and relevant metadata from different sources, and organizes them into parallelizable shards for subsequent batch teacher inference. Unlike cleaned datasets constructed in advance for a particular teacher model, this work does not introduce teacher-specific filtering rules during candidate-pool construction. This choice preserves the realistic heterogeneity of the public candidate structure pool as much as possible and reduces prior bias in subsequent teacher evaluation and routing learning.

\subsection{Construction of the \texorpdfstring{r\textsuperscript{2}SCAN}{r2SCAN} calibration set}
The real r\textsuperscript{2}SCAN calibration set is sampled and computed from the large-scale candidate structure pool. It contains approximately $1.8\times10^{4}$ structures and is used for candidate teacher evaluation, supervised-label construction and ATR router training. This set is not derived from a predefined benchmark. Instead, it is sampled directly from the candidate structure pool used in subsequent large-scale pseudo-label deployment, making it more suitable as target-distribution calibration data. The sampling process follows trajectory-aware, element-coverage-aware and structure-diversity-aware principles. Specifically, representative structures are preferentially selected from each trajectory to reduce evaluation bias caused by adjacent trajectory frames and near-duplicate configurations. At the same time, element coverage, chemical-system coverage and structural-type diversity are expanded as much as possible during sampling. Detailed composition statistics of this calibration set, including element coverage, chemical-system distribution, frame duplication and structural-size distribution, are provided in Results 2.2. To avoid information leakage, all training, validation, test and cross-validation procedures based on this calibration set are grouped at the structure level, rather than by randomly splitting individual structure--teacher pairs. This ensures that all teacher-prediction samples corresponding to the same structure or material group appear only in the same split, avoiding optimistic evaluation caused by near-duplicate structures, shared structural features or multi-teacher predictions.

\subsection{Adaptive Multi-Teacher Routing (ATR): feature construction, label definition, and deployment policy}
To construct high-quality r\textsuperscript{2}SCAN-level pseudo-labels on the large-scale candidate structure pool, this work proposes the ATR framework. ATR does not learn continuous error regression. Instead, it performs structure-wise acceptability classification for each structure--teacher pair, followed by unique teacher selection or rejection after multi-teacher predictions are available. If at least one teacher satisfies the predefined quality criterion, ATR outputs the highest-scoring teacher and its label. If none of the candidate teachers meets the requirement, the structure enters the reject pool. The basic training unit of ATR is therefore not an individual structure, but a structure--teacher pair $(x_i,M_k)$. Real r\textsuperscript{2}SCAN DFT labels are used only offline to construct supervised labels $y_{ik}$, and are not used as input features during ATR deployment. When ATR is applied to large-scale unlabeled structures, the model relies only on structural descriptors, teacher predictions and inter-teacher disagreement features. The router learns whether teacher $M_k$ is sufficiently reliable on structure $x_i$, rather than directly performing single-label classification for the structure as a whole.

The final deployment version in this work uses a binary router, whose general form is
\begin{equation}
\mathrm{ATR}=\mathrm{SimpleImputer}(\mathrm{median})+g_\theta,
\end{equation}
where missing features are first imputed using the median of the training-set features, and $g_\theta$ denotes a lightweight tabular binary classifier that can be calibrated. In the main results of this work, $g_\theta$ is implemented using XGBoost. We also compare RandomForest and LightGBM, finding that similar tree-model backends give the same trends. The final deployment teacher set is denoted by $\mathcal{T}=\{M_k\}_{k=1}^{K}$, where $K=5$.

For the $i$-th structure $x_i$ and the $k$-th candidate teacher $M_k$, let the energy and force predictions given by this teacher be $E_i^{(k)}$ and $\mathbf{F}_{i,j}^{(k)}$, respectively, and let the corresponding r\textsuperscript{2}SCAN single-point DFT reference labels be $E_i^{\mathrm{DFT}}$ and $\mathbf{F}_{i,j}^{\mathrm{DFT}}$. If structure $x_i$ contains $N_i$ atoms, the per-atom energy error and maximum atomic force error are defined in this work as
\begin{equation}
\Delta E_{ik}=\frac{\left|E_i^{(k)}-E_i^{\mathrm{DFT}}\right|}{N_i},
\end{equation}
\begin{equation}
\Delta F^{\max}_{ik}=\max_j\left\|\mathbf{F}_{i,j}^{(k)}-\mathbf{F}_{i,j}^{\mathrm{DFT}}\right\|_2,
\end{equation}
where $j$ runs over all atoms in the structure.

The final deployment in the main text adopts a binary A/R label scheme. If a given $(x_i,M_k)$ simultaneously satisfies the thresholds for per-atom energy error and maximum atomic force error, it is labeled Accept; otherwise, it is labeled Reject:
\begin{equation}
y_{ik}=\begin{cases}
1, & \Delta E_{ik}<0.1\ \mathrm{eV/atom}\ \mathrm{and}\ \Delta F^{\max}_{ik}<1.0\ \mathrm{eV}/\mathring{\mathrm{A}},\\
0, & \mathrm{otherwise}.
\end{cases}
\end{equation}
Here $y_{ik}=1$ means that the structure--teacher pair satisfies the Accept criterion, and $y_{ik}=0$ means Reject. Because the same structure corresponds to multiple $(x_i,M_k)$ samples, all training, validation and test splits are grouped at the structure/immutable-id level. This ensures that all teacher samples corresponding to the same structure appear only in the same split, avoiding information leakage caused by different teacher predictions sharing the same structure. Thus, ATR effectively learns a conditional probability model for estimating whether a given teacher satisfies the Accept criterion on a given structure:
\begin{equation}
p_{ik}=P(y_{ik}=1\mid \mathbf{z}_{ik}),
\end{equation}
where $\mathbf{z}_{ik}$ denotes the input features associated with structure $x_i$ and teacher $M_k$.

The final input features can be written as
\begin{equation}
\mathbf{z}_{ik}=\left[\mathbf{s}_i,\mathbf{m}_k,\mathbf{r}_{ik}\right].
\end{equation}
In the final deployment bundle of this work, the actual input dimensionality passed to the router is 216, which can be decomposed into 177 structure features, 5 model features and 34 router features. The structure features describe geometry and chemical composition, including 45 basic structural descriptors, 118 element-fraction channels, 1 OOD feature and 13 clustering-related features. The basic descriptors include the number of atoms, density, lattice parameters, volume per atom, nearest-neighbor distance statistics and coordination statistics. The element-fraction channels are denoted as \texttt{structure\_\_elem\_frac\_z***}, and the OOD feature is \texttt{structure\_\_ood\_knn\_mean}. The model features represent the identity of the current candidate teacher in one-hot form. The router features encode local behavioral differences between the current teacher and the top-5 strong-teacher set. These include the deviation of the current teacher's energy from the teacher-set median, inter-teacher energy standard deviation, force-component-level median deviation, maximum force-field disagreement, near-zero-force fraction and compressed anomaly-proxy features. Representative energy-disagreement features can be written as
\begin{equation}
\mathrm{std}_E(i)=\mathrm{Std}\left(\left\{E_i^{(k)}/N_i\right\}_{k\in\mathcal{T}}\right),
\end{equation}
\begin{equation}
\mathrm{med}_E(i)=\mathrm{Median}\left(\left\{E_i^{(k)}/N_i\right\}_{k\in\mathcal{T}}\right),
\end{equation}
\begin{equation}
\delta E_{ik}=\left|E_i^{(k)}/N_i-\mathrm{med}_E(i)\right|.
\end{equation}
Similarly, statistics such as inter-teacher mean, standard deviation, median deviation, maximum deviation, near-zero fraction and local extreme-component fraction are constructed for forces. If multiple strong teachers are highly consistent on a structure, the structure is more likely to be a high-confidence sample. Conversely, if teacher disagreement is substantial, the structure is more likely to lie in a hard-sample or out-of-distribution region.

At inference time, for multiple teacher candidates associated with the same structure $x_i$, ATR first compares their acceptance probabilities and then performs final routing according to the threshold $\tau$:
\begin{equation}
k^*=\arg\max_{k\in\mathcal{T}}p_{ik}.
\end{equation}
If
\begin{equation}
\max_{k\in\mathcal{T}}p_{ik}\ge \tau,
\end{equation}
then the prediction of teacher $M_{k^*}$ is accepted as the pseudo-label for the structure; otherwise, the structure is assigned to the reject pool. Equivalently, the final structure-wise output is
\begin{equation}
\hat{y}_i=\begin{cases}
M_{k^*}, & \max_{k\in\mathcal{T}}p_{ik}\ge \tau,\\
\mathrm{Reject}, & \max_{k\in\mathcal{T}}p_{ik}<\tau.
\end{cases}
\end{equation}
The threshold $\tau$ is selected only on the validation/calibration split corresponding to the training fold, in order to satisfy a predefined precision target and control the coverage--precision trade-off. Once determined, it is fixed for the corresponding test set or large-scale deployment, avoiding post hoc adjustment of the threshold on test results.

The final practical version further applies lightweight postprocessing on top of the main routing rule to suppress local anomalous hard cases. This postprocessing relies only on teacher-only disagreement features and does not use real DFT labels. Specifically, when the acceptance probability of the selected teacher is below 0.45 and its maximum force-field disagreement proxy relative to the strong-teacher set is greater than 0.15, the structure is secondarily assigned to the reject pool. Therefore, ATR is not a fixed-error-threshold filter. It is a structure-wise routing policy calibrated using a small amount of real r\textsuperscript{2}SCAN annotations, enabling teacher selection, risk-aware rejection and pseudo-label generation on large-scale unlabeled structures.

\subsection{UMAP visualization and dataset distribution analysis}
UMAP visualization is used to qualitatively compare the relative positions of the ATR high-confidence subset, existing MP-r\textsuperscript{2}SCAN data, the held-out DFT test set and the MD validation systems in a common composition--structure feature space\cite{McInnes2018UMAP}. The Full candidate pool, containing $112,932,152$ structures, defines the global materials-space reference for the embedding, but Full points are not displayed in the final figure to avoid visual overcrowding. The ATR high-confidence Subset contains $2,893,331$ structures, and the MP-r\textsuperscript{2}SCAN dataset contains $190,597$ structures. Coverage-preserving samples corresponding to approximately 10\% of Subset and MP-r\textsuperscript{2}SCAN are displayed, together with all 2,000 held-out r\textsuperscript{2}SCAN DFT test structures. The three MD systems HgInSn, InSe and MgSi are marked with special red symbols. This visualization is not used for ATR training or the final screening decision; it is used only to interpret dataset relationships and the selection of validation systems. Because the plotted datasets are sampled differently, point density should be interpreted only as relative distributional position, not as an absolute probability density in the original data space.

\subsection{Model pretraining and downstream evaluation}
To verify whether the r\textsuperscript{2}SCAN-level pseudo-labels constructed by ATR can be converted into actual model-training benefits, this work uses CHGNet as the student model for pretraining and downstream fine-tuning experiments\cite{Deng2023CHGNet}. The pretraining stage uses approximately $2.6\times10^{6}$ ATR-accepted structures. The model adopts a randomly initialized CHGNet architecture and learns both energies and forces. The training loss uses the Huber loss, written as
\begin{equation}
\mathcal{L}=w_E\mathcal{L}_E+w_F\mathcal{L}_F,
\end{equation}
where $w_E:w_F=1:1$ during pretraining. Pretraining is performed with 4-GPU DDP training, with a per-GPU batch size of 64 and a global batch size of 256. The initial learning rate is $1\times10^{-3}$, and the learning-rate scheduler uses CosineAnnealingLR for cosine decay, with a minimum learning rate of $1\times10^{-5}$. The total number of training epochs is 40. The best pretraining checkpoint is selected according to validation-set performance.

Subsequently, the student model is fine-tuned on the MP-r\textsuperscript{2}SCAN dataset. In the fine-tuning stage, the best checkpoint obtained from pseudo-label pretraining is used as initialization. Huber loss is still used, but the loss weights of energy and force are adjusted to $w_E:w_F=3:1$ to strengthen the model's ability to fit the energy scale. Fine-tuning is also performed with 4-GPU DDP training, with a per-GPU batch size of 64 and a global batch size of 256. The initial learning rate is $1\times10^{-3}$ and is decayed by CosineAnnealingLR to a minimum learning rate of $1\times10^{-5}$. The total training length is 50 epochs. The best model appears around epoch 35 and is selected according to validation metrics for subsequent evaluation.

To distinguish the contributions of pseudo-label quality, the training workflow and data volume, this work sets up control groups. First, the baseline uses the same CHGNet architecture but does not perform ATR pseudo-label pretraining, and is trained only on the MP-r\textsuperscript{2}SCAN dataset. Second, the without ATR controls perform supervised pretraining on the original labels of the same structures, as well as self-supervised and supervised pretraining followed by MP-r\textsuperscript{2}SCAN fine-tuning. These controls test whether more pretraining structures or training steps alone can explain the improvement from ATR. Their performance in the present tests indicates that simply increasing the number of training structures and training steps is insufficient to replace high-confidence r\textsuperscript{2}SCAN-level pseudo-labels screened through ATR routing.

Downstream evaluation includes two types of tests. One evaluates model extrapolation performance on 2,000 held-out r\textsuperscript{2}SCAN DFT structures that do not overlap with the pseudo-label training set. The other evaluates model adaptability to the standard r\textsuperscript{2}SCAN data distribution on the public MP-r\textsuperscript{2}SCAN benchmark. All evaluations report energy MAE and force MAE, with energy error reported in eV/atom and force error in eV/\AA{}. To avoid data leakage, pseudo-label subset splitting, pretraining validation-set construction and subsequent held-out testing are all grouped at the trajectory or immutable-id level. This ensures that configurations from the same source do not appear repeatedly across training, validation and test sets. Before evaluation, the 2,000 held-out r\textsuperscript{2}SCAN DFT test structures are also checked for duplicates against the ATR pseudo-label pretraining set, the MP-r\textsuperscript{2}SCAN fine-tuning set and the MP-r\textsuperscript{2}SCAN benchmark.

\subsection{\texorpdfstring{r\textsuperscript{2}SCAN}{r2SCAN} DFT calculations and AIMD/MLMD validation protocol}

All real r\textsuperscript{2}SCAN labels in this work are obtained using VASP, including static single-point labels for calibration/test structures and AIMD reference trajectories. For the single-point data required by the calibration set, held-out test set and router training, we use a MatPES-compatible static VASP workflow to obtain total energies, atomic forces, stress tensors and magnetic moments\cite{Kaplan2025MatPES}. The exchange--correlation functional is the r\textsuperscript{2}SCAN meta-GGA, set as \texttt{METAGGA = R2SCAN}, and PBE\_64 PAW potentials are used\cite{Furness2020R2SCAN,Kingsbury2022R2SCAN}. The plane-wave cutoff energy is set to \texttt{ENCUT = 680 eV}, the augmentation-grid cutoff energy is set to \texttt{ENAUG = 1360 eV}, and the calculation precision is \texttt{PREC = Accurate}. Brillouin-zone sampling is generated automatically by VASP using \texttt{KSPACING} = $0.22~\mathring{\mathrm{A}}^{-1}$. All single-point calculations use \texttt{NSW = 0}, and the electronic self-consistency convergence criterion is \texttt{EDIFF = 1E-5}. To preserve possible magnetic information, spin-polarized settings are used, namely \texttt{ISPIN = 2}, and site-projected magnetic moments are output through \texttt{LORBIT = 11}. Finally, zero-smearing extrapolated total energies, Hellmann--Feynman forces, symmetric stress tensors and magnetic moments are extracted from the VASP output. The current ATR label definition and student-model training mainly use total energies and atomic forces. Stress tensors and magnetic moments are retained during the calculations, but they are not included in the current main workflow and are kept as a basis for future extensions to more complete physical constraints.

To test the stability of the ATR student model in finite-temperature dynamics, this work further selects three systems, Hg$_{16}$In$_{20}$Sn$_{12}$, Mg$_{24}$Si$_{20}$ and In$_{24}$Se$_{24}$, to construct 300 K r\textsuperscript{2}SCAN AIMD reference data. AIMD is also performed using VASP with the r\textsuperscript{2}SCAN meta-GGA functional\cite{Furness2020R2SCAN,Kaplan2025MatPES}. All three systems use periodic supercells. The plane-wave cutoff energy is set to 450 eV, and $k$-point sampling uses a Gamma-only setting. The calculations use \texttt{PREC = Accurate} and \texttt{LASPH = .TRUE.}, and set \texttt{ISYM = 0} to avoid symmetry constraints affecting atomic motion during molecular dynamics. HgInSn and MgSi use metallic electronic smearing settings, \texttt{ISMEAR = 1} and \texttt{SIGMA = 0.10 eV}; InSe uses \texttt{ISMEAR = 0} and \texttt{SIGMA = 0.05 eV}.

All AIMD simulations are performed in the NVT ensemble with a time step of 1.0 fs and a target temperature of 300 K. HgInSn completes a 12.5 ps r\textsuperscript{2}SCAN NVT AIMD reference trajectory with 12,500 frames. The first approximately 2.5 ps are used for thermalization and stable MLMD initial-structure selection, and the subsequent 10 ps are used for formal DFT-MD and model comparison. MgSi and InSe each complete a 2 ps r\textsuperscript{2}SCAN NVT AIMD reference trajectory, each containing 2,000 frames. For all trajectories, cells, atomic positions, total energies and atomic forces are saved and exported in extxyz format for model evaluation.

In MLMD validation, we select stable intermediate frames from the DFT-MD trajectories as initial structures and perform 300 K NVT MLMD simulations using the student model and the corresponding baseline model. For the HgInSn system, CHGNet-r\textsuperscript{2}SCAN\cite{Huang2025CrossFunctional}, SevenNet-Omni-i12-MP and SevenNet-Omni-i12-MatPES\cite{Kim2026SevenNetOmni} are further included as strong teacher baselines. MLMD stability is diagnosed using temperature, potential energy, kinetic energy, maximum atomic force, nearest-neighbor distance and partial RDF. For the HgInSn system, energy MAE is computed on the 10 ps DFT-MD reference trajectory for quantitative evaluation of each model, whereas the other systems are evaluated qualitatively.

\section{Conclusions}

This work proposes the Adaptive Multi-Teacher Routing (ATR) framework for constructing high-quality r\textsuperscript{2}SCAN-level pseudo-labels from large-scale public candidate structure pools. By combining multi-teacher complementarity, structure-wise risk control and an explicit rejection mechanism, ATR turns teacher selection and high-risk pseudo-label rejection into a learnable and verifiable decision process. The experimental results show that ATR achieves a stable balance between coverage and precision, reduces the risk of false acceptance caused by uniform single-teacher labeling, and provides a more reliable basis for large-scale high-accuracy pseudo-label data construction. Student-model validation further shows that ATR pseudo-labels improve static DFT test errors and enhance dynamical stability in 300 K MLMD for the three systems HgInSn, MgSi and InSe.

On this basis, this work further demonstrates a closed-loop system for practical deployment. The workflow starts from large-scale candidate structure collection from public materials databases, uses small-scale real r\textsuperscript{2}SCAN annotations for calibration, and applies ATR screening to form a high-confidence pseudo-label dataset for student-model training. The results show that this closed loop can substantially reduce the cost of constructing large-scale high-accuracy materials data without requiring full r\textsuperscript{2}SCAN recomputation of the candidate pool. It also improves student-model generalization on r\textsuperscript{2}SCAN-level prediction tasks and yields a more robust potential-energy surface in finite-temperature dynamical tests. In the future, ATR can be extended to richer physical constraints such as stress, magmom, trajectory consistency and relaxation stability. It can also incorporate broader teacher-model families, higher-fidelity label systems and cross-domain atomistic simulation scenarios, including molecules, surfaces and interfaces. We hope that ATR can serve as a step toward the continuous expansion of high-fidelity materials data, promote more efficient construction and sharing of high-quality training data, and further advance data-driven materials discovery.

\bibliographystyle{unsrt}
\bibliography{references_0701}
\end{document}